\pdfoutput=1
\documentclass[10pt, a4paper]{article}
\usepackage{lrec}
\usepackage{titlesec}
\titleformat{\section}{\normalfont\large\bf\center}{\thesection.}{1em}{}
\titleformat{\subsection}{\normalfont\SmallTitleFont\bf\raggedright}{\thesubsection.}{1em}{}
\titleformat{\subsubsection}{\normalfont\normalsize\bf\raggedright}{\thesubsubsection.}{1em}{}
\renewcommand\thesection{\arabic{section}}
\renewcommand\thesubsection{\thesection.\arabic{subsection}}
\renewcommand\thesubsubsection{\thesubsection.\arabic{subsubsection}}
\usepackage{graphicx}
\usepackage{tabularx}
\usepackage{soul}
\usepackage{subcaption}
\usepackage{epstopdf}
\usepackage[T1]{fontenc}
\usepackage[utf8]{inputenc}
\usepackage{hyperref}
\usepackage{xstring}

\usepackage{tikz-qtree}
\usepackage{tikz-dependency}%% For dependency graphs≈ß

\newcommand\tag[1]{\textsc{#1}}

\tikzset{edge from parent/.style={->,draw,font={\scshape\small}},
    every tree node/.style={align=center,anchor=north},
    level distance=10ex}

\usepackage{covington}
\usepackage{mathptmx}
\usepackage{url}
\usepackage{latexsym}
\usepackage{booktabs}
\usepackage{multirow}
\usepackage{todonotes}

\newcommand{\etype}[1]{\texttt{#1}}
\newcommand{\norex}[1]{\textit{#1}}
\newcommand{\eng}[1]{`#1'}
\newcommand{\nob}[0]{BM}
\newcommand{\nno}[0]{NN}

\title{NorNE: Annotating Named Entities for Norwegian}

\name{Fredrik Jørgensen,$^\dagger$ Tobias Aasmoe,$^\ddagger$ Anne-Stine Ruud Husevåg,$^\diamondsuit$ Lilja Øvrelid,$^\ddagger$ Erik Velldal$^\ddagger$}

\address{Schibsted Media Group,$^\dagger$ Oslo Metropolitan University,$^\diamondsuit$ University of Oslo$^\ddagger$\\
         \texttt{fredrik.jorgensen@schibsted.com},$^\dagger$  \texttt{annesh@oslomet.no},$^\diamondsuit$\\ 
         \texttt{\{tobiaaa,liljao,erikve\}@ifi.uio.no}$^\ddagger$\\}

\abstract{
This paper presents NorNE, a manually annotated corpus of named entities which extends the annotation of the existing Norwegian Dependency Treebank. Comprising both of the official standards of written Norwegian (Bokmål and Nynorsk), the corpus contains around 600,000 tokens and annotates a rich set of entity types including persons, organizations, locations, geo-political entities, products, and events, in addition to a class corresponding to nominals derived from names. We here present details on the annotation effort, guidelines, inter-annotator agreement and an experimental analysis of the corpus using a neural sequence labeling architecture.
 \\ \newline 
 }

\begin{document}

\maketitleabstract

\section{Introduction}
\label{sec:intro}

This paper documents the efforts  
of creating the first publicly available dataset for named entity recognition (NER) for Norwegian, dubbed NorNE.\footnote{\url{https://github.com/ltgoslo/norne/}}  The dataset adds named entity annotations on top of the Norwegian Dependency Treebank (NDT) \cite{Sol:Skj:Ovr:14}, containing  manually  annotated  syntactic and morphological information for both varieties of written Norwegian -- Bokmål and Nynorsk -- comprising roughly 300,000 tokens of each. The corpus contains mostly news texts (around 85\% of the corpus), but also other types of texts, such as government reports, parliament transcripts and blogs. The treebank has following its release also been converted to the Universal Dependencies standard \cite{Ovr:Hoh:16,VelOvrHoh17}. We correspondingly distribute the annotations of NorNE in two versions, mirroring both the original NDT and the UD-version.

The annotations in NorNE include a rich set of entity types. In short, they comprise the following (more details are given in the guideline discussion in Section~\ref{sec:annotation}): 
\begin{list}{$\circ$}{} 
    \item \textbf{Person} (\etype{PER}): Named real or fictional characters. 
    \item \textbf{Organization} (\etype{ORG}): Any collection of people, such as firms,  music groups, 
    political parties etc.
    \item \textbf{Location} (\etype{LOC}): Geographical places 
    and facilities. 
    \item \textbf{Geo-political entity} (\etype{GPE}): Geographical regions defined by political and/or social groups,  
    additionally sub-categorized as either: 
    \begin{itemize}
     \item GPE with a locative sense (\etype{GPE\_LOC}),
    \item GPE with an organization sense (\etype{GPE\_ORG}).
    \end{itemize}
    \item \textbf{Product} (\etype{PROD}): Artificially produced entities, including abstract entities such as radio shows, programming languages, ideas, etc.
   \item \textbf{Event} (\etype{EVT}): Festivals, weather phenomena, etc. 
   \item \textbf{Derived} (\etype{DRV}): Nominals 
   that are derived from a name, but not a named entity in themselves.
\end{list}

In addition to discussing the annotation process and guidelines, we also provide an exploratory analysis of the resulting dataset through a series of experiments using state-of-the-art neural architectures for named entity recognition (combining a character-level CNN and a word-level BiLSTM, feeding into a CRF inference layer). The purpose of the experimental section is to provide some preliminary baseline results while simultaneously validating the consistency and usability of the annotations, and finally also shedding some light on the consequences of different design choices that can be made in the modeling stage, like the choice of which label set to use or which label encoding to use and so on.  

The remainder of the paper is organized as follows. First, 
Section~\ref{sec:previous} briefly outlines previous work on NER for Norwegian. Section~\ref{sec:annotation} then describes the NorNE annotation effort in more detail, briefly outlining the annotation guidelines while providing illustrating examples and presenting an analysis of inter-annotator agreement. In Section~\ref{sec:dataset} we summarize the resulting dataset before we in Section~\ref{sec:experiments} turn to an empirically-driven analysis of the data through a series of experiments. 

\section{Previous Work on NER for Norwegian}
\label{sec:previous}

While NorNE is the first publicly available dataset for NER in Norwegian, there have been some previous efforts to create similar resources.  

Most notably, in a parallel effort, and as part of his doctoral work, \newcite{Johansen:19a} added named entity annotations to the same underlying corpus, the UD-version of NDT. However, the data was single-annotated and using a reduced label-set with only 4 different categories -- locations, organizations, persons, and miscellaneous -- and only considering text spans already tagged as PROPN (proper noun) in the original treebank data.

Previously, the Nomen Nescio project \cite{Joh:Hag:Haa:05}, focusing on Scandinavian languages, created a named entity annotated corpus for Norwegian. However, due to copyright restrictions the data was unfortunately not made publicly available. The Nomen Nescio corpus was based on articles from several news papers and magazines, in addition to some works of fiction. It totaled 226,984 tokens, of which 7,590 were part of a named entity annotation, and the following six entity categories were used: person, organization, location, event, work of art, and miscellaneous. For a while, Norwegian named entity recognition based on these categories was supported in the Oslo--Bergen tagger \cite{Joh:Hag:Lyn:12}, a rule-based morpho-syntactic tagger and lemmatizer for Norwegian based on constraint grammar, but unfortunately NER is no longer supported in the tagger. For more information on the Nomen Nescio data, including experimental results, see \newcite{Noklestad:09}.

The TORCH (Transforming the Organization and Retrieval of Cultural Heritage) project \cite{Hof:Pre:15,Tal:Mas:Hus:14} focused on  automatic metadata generation to improve access to digitized cultural expressions. As part of this effort, a corpus of subtitles and metadata descriptions from the archive of NRK (the national public broadcaster), was annotated with named entities. To facilitate comparison to previous work on Norwegian NER, the project adopted the categories used in Nomen Nescio, also extending the original annotation guidelines \cite{Jon:03}, as documented in \cite{How:14}. 
However, as some of the data contain confidential information and was only made available to the researchers on the basis of a non-disclosure agreement, the resulting dataset could unfortunately not be made publicly available.

\section{Annotation}
\label{sec:annotation}

In this section we discuss in more detail various aspects related to the annotations; we first describe some relevant properties of named entities in Norwegian, and go on to flesh out the entity types in more detail, we highlight important parts of the guidelines, discuss issues that are particular to mentions of named entities in Norwegian and finally present an analysis of inter-annotator agreement. 

\subsection{Proper Nouns in Norwegian}
\label{sec:nes}
In Norwegian, proper nouns are generally capitalized. For multi-token names, however, the general rule is that only the first token should be capitalized \cite{Faa:Lie:Van:97}, e.g. \norex{Oslo rådhus} \eng{Oslo city hall}. There are a number of exceptions to this general rule as well, for instance in company names, e.g. \norex{Den Norske Bank} \eng{The Norwegian Bank}. Proper nouns may also be nested to form a larger unit, e.g. \norex{Universitetet i Oslo} \eng{University of Oslo}.
Unlike in English, names for days, months and holidays are not capitalized, e.g. \norex{onsdag} \eng{Wednesday}, \norex{juli} \eng{July}.
Like most Germanic languages, compounding is highly productive in Norwegian and compounds are written as one token, e.g. \norex{designbutikk} \eng{designer store}. However, when the initial part of a compound is a proper name, these are separated by a hyphen, e.g. \norex{Prada-butikken} \eng{the Prada store}.

\subsection{Entity Types in NorNE} 

In the NorNE corpus we annotate the six main types of named entities \etype{PER}, \etype{ORG}, \etype{LOC}, \etype{GPE}, \etype{PROD} and \etype{EVENT}, as mentioned in Section~\ref{sec:intro} above (in addition to a special category for nominals \textit{derived} from proper names). 
In the following we will present the different categories and the main principles behind their annotation. We will comment on how various aspects of the annotation relates to other well-known datasets, but also discuss entity properties that are specific to Norwegian.

\paragraph{Person (\etype{PER})}
The person name category includes names of real people and fictional characters. Names of other animate beings such as animals are also annotated as \etype{PER}, e.g., \norex{Lassie} in \norex{den kjente TV-hunden Lassie} \eng{the famous TV-dog Lassie}. Family names should be annotated as \etype{PER} even though they refer to several people. 

\paragraph{Organization (\etype{ORG})} This entity category includes any named group of people, such as firms, institutions, organizations, pop groups, political parties etc. \etype{ORG} also includes names of places when they act as administrative entities, as in (\ref{ex:org1}) below which annotates sport teams associated with a location. 
Corporate designators like \norex{AS}, \norex{Co.} and \norex{Ltd.} should always be included as part of the named entity, as in (\ref{ex:org2}).
\begin{examples}
\item\label{ex:org1}
\gll \underline{V{\aa}lerenga}$_{ORG}$ tapte mot \underline{Troms{\o}$_{ORG}$}.
V{\aa}lerenga lost against Troms{\o}
\glt `V{\aa}lerenga lost against Troms{\o}'
\glend
\item\label{ex:org2}
\gll Advokatfirmaet \underline{Lie \& Co}$_{ORG}$ representerer \underline{Hansen}$_{PER}$.
Lawyers {Lie \& Co} represent Hansen
\glt `The lawyers Lie \& Co represent Hansen'
\glend
\end{examples}

\paragraph{Location (\etype{LOC})} This entity category denotes geographical places, buildings and various facilities. Examples are airports, churches, restaurants, hotels, hospitals, shops, street addresses, roads, oceans, fjords, mountains and parks.
Postal addresses are not annotated, but the building, town, county and country within the address are to be annotated, all as LOC entities, see (\ref{ex:loc1}).
\begin{examples}
\item\label{ex:loc1}
\gll \underline{{\O}vregaten 2a}$_{LOC}$, 5003 \underline{Bergen}$_{LOC}$
{{\O}vre-street 2a}, 5003 Bergen
\glt `{\O}vre-street 2a, 5003 Bergen'
\glend
\end{examples}

\paragraph{Geo-political entity (\etype{GPE})} 
Similarly to OntoNotes \cite{Wei:Pal:Mar:13}, but also the Groningen Meaning Bank (GMB) \cite{Bos:Bas:Eva:17}, we annotate \textit{geo-political entities} (GPEs). This category was introduced in the annotation scheme of the Automatic Content Extraction program (ACE) \cite{Mit:Str:Prz:03}, and GPEs will further have either location or organization as its sub-type (these are dubbed `mention roles' in ACE, where also additional such sub-types are defined). 

Following \newcite{Mit:Str:Prz:03}, 
\etype{GPE} entities denote geographical regions that are defined by political and/or social groups.  
GPES are describe complex entities that refer both to a population, a government, a location, and possibly also a nation (or province, state, county, city, etc.). A \etype{GPE} must be one of the following: a nation, city or region with a parliament-like government. Parts of cities and roads are not annotated as \etype{GPE}s.

As mentioned above, \etype{GPE} entities are further subtyped, either as \etype{GPE\_LOC} or \etype{GPE\_ORG}. If a sense is mostly locative, it should be annotated as \etype{GPE\_LOC}, otherwise it should be \etype{GPE\_ORG}. Example (\ref{ex:gpe1}) below shows both these entity types.
\begin{examples}
\item\label{ex:gpe1}
\gll \underline{Norge}$_{GPE\_ORG}$ reagerer p{\aa} politivolden i \underline{Catalonia}$_{GPE\_LOC}$
Norway reacts to {police violence} in Catalonia
\glt `Norway reacts to the police violence in Catalonia'
\glend
\end{examples}
Sometimes the names of \etype{GPE} entities may be used to refer to other referents  associated with a region besides the government, people, or aggregate contents of the region. The most common examples are sports teams. 
In our annotation scheme, these entities are marked as teams (\etype{ORG}), as they do not refer to any geo-political aspect of the entity. 

\paragraph{Product (\etype{PROD})} In line with other recent named entity corpora like OntoNotes 5.0 \cite{Wei:Pal:Mar:13} and MEANTIME \cite{Min:Spe:Uri:16}, we also annotate \textit{products} as a separate category. Note that while OntoNotes additionally label \textit{works-of-art} as a separate category, as was also done in the Norwegian Nomen Nescio corpus \cite{Joh:Hag:Haa:05}, this is subsumed by the product category in NorNE.

All entities that refer to artificially produced objects are annotated as products. This includes more abstract entities, such as speeches, radio shows, programming languages, contracts, laws and even ideas (if they are named). Brands are \etype{PROD} when they refer to a product or a line of products, as in (\ref{ex:prod1}), but \etype{ORG} when they refer to the acting or producing entity (\ref{ex:prod2}):
\begin{examples}
\item\label{ex:prod1}
\gll \underline{Audi}$_{PROD}$ er den beste bilen.
Audi is the best car.
\glt `Audi is the best car'
\glend
\item\label{ex:prod2}
\gll \underline{Audi}$_{ORG}$ lager de beste bilene.
Audi makes the best cars.
\glt `Audi makes the best cars'
\glend
\end{examples}

\paragraph{Event (\etype{EVT})}  
 Again similarly to OntoNotes \cite{Wei:Pal:Mar:13}, but also the Groningen Meaning Bank (GMB) \cite{Bos:Bas:Eva:17}, NorNE also annotates  \textit{events}. 
This category includes names of festivals, cultural events, sports events, weather phenomena and wars. Events always have a time span, and often a location where they take place as well.
An event and the organization that arranges the event can share a name, but should be annotated with different categories: \etype{ORG}, as in (\ref{ex:event1}) vs. \etype{EVT}, as in (\ref{ex:event2}).
\begin{examples}
\item\label{ex:event1}
\gll \underline{Quartfestivalen}$_{ORG}$ gikk konkurs i 2008.
Quart-festival went bankrupt in 2008.
\glt `The Quart Festival went bankrupt in 2008'
\glend
\item\label{ex:event2}
\gll \underline{Rolling Stones}$_{ORG}$ fikk dessverre aldri spilt p{\aa} \underline{Quartfestivalen}$_{EVT}$.
{Rolling Stones} got unfortunately never played at Quart-festival
\glt `The Rolling Stones unfortunately never got to play at the Quart Festival'
\glend
\end{examples}

\paragraph{Derived (\etype{DRV})} The \etype{DRV category} is a special category for compound nominals that contain a proper name, as described in section \ref{sec:nes} above. The main criteria for this category is that the entity in question (i) contains a full name, (ii) is  capitalized, and, (iii) is not itself a name. Examples include for instance \norex{Oslo-mannen} \eng{the Oslo-man}.
Names that are inflected and used as common nouns, are also tagged derived (DRV), e.g. \norex{Nobel-prisene} \eng{the Nobel-prizes}.
The reason for the special treatment of these types of forms is that these words do not have a unique entity as a reference, but rather exploit an entity as part of their semantics. Even so, if a user of the annotated data wishes to extract all information about a particular named entity, these may still be relevant, hence should be marked separately.

\paragraph{Entity types not included} 
Of the categories discussed above, \textit{location}, \textit{person}, and \textit{organization} comprise the core inventory of named entity types in the literature. They formed part of the pioneering shared tasks on NER hosted by CoNLL 2002/2003 \cite{Sang:02,San:Meu:03}, MUC-6 \cite{Gri:Sun:95} and MUC-7 \cite{Chinchor:98}, and have been part of all major NER annotation efforts since.  
However, the CoNLL shared tasks also included a fourth category for names of \textit{miscellaneous} entities not belonging to the aforementioned three. During the annotation of NorNE we similarly operated with an entity type \etype{MISC}, but eventually we decided to discard this label in the final release of the data as it was annotated too rarely to be useful in practice (with a total of 8 occurrences in the training data but 0 occurrences in both the development and held-out splits).

The MUC shared tasks, on the other hand, additionally included identification of certain types of \textit{temporal expressions} (date and time) and \textit{number expressions} (monetary expressions and percentages). We did not include these in NorNE as we do not strictly speaking consider them named entities. 

\subsection{Annotation Guidelines}
\label{sec:guidelines}

We now turn to present the most relevant aspects of the annotation guidelines developed in the project. Note that the complete annotation guidelines are  distributed with the corpus. The guidelines were partly based on \newcite{Jon:03} and \newcite{How:14}, as well as guidelines created for English corpora, in particular the ACE \cite{Mit:Str:Prz:03} and CoNLL \cite{San:Meu:03} datasets. 

\subsubsection{Main Annotation Criteria}
The guidelines formulate a number of criteria for the annotations. In general, the annotated entities should have a unique reference which is constant over time. Further, the annotated text spans should adhere to the following annotation criteria:
\paragraph{Proper nouns} The text span corresponds to or contains a proper noun, e.g. \norex{Per Hansen} which consists of two consecutive proper nouns with a single reference. Names may include words which are not proper nouns, e.g. prepositions, as in \norex{Universitetet i Oslo} \eng{University of Oslo}.

\paragraph{Span}
The maximum span of a name should be annotated, rather than its parts, e.g. in \norex{H{\o}gskolen i Oslo og Akershus} \eng{the College of Oslo and Akershus}, \norex{Oslo} and \norex{Akershus} are not annotated separately as locations.

\paragraph{Capitalization} 
As mentioned above, it is often the case in Norwegian that only the first token in a multi-token named entity is capitalized. We treat the subsequent tokens as part of the name if they in combination denote a unique named entity, e.g. \norex{Oslo r{\aa}dhus} \eng{Oslo city-hall}. 

\paragraph{Titles}
Most titles do not have an initial capital letter in Norwegian. Exceptions are some instances of royal titles. We never annotate titles as a name or part of a name, even when they are capitalized.
Official work titles like \norex{Fylkesmannen} `the County official' should be annotated as organizations because they refer to official institutions, as in (\ref{ex:title1}). When the same entity refers to the person/occupation, as in (\ref{ex:title2}), they are not, however, annotated as named entities.

\begin{examples}
\item\label{ex:title1}
\gll I g{\aa}r ble det klart at \underline{Fylkesmannen}$_{ORG}$ har vedtatt \ldots
in yesterday was it clear that County-man has declared \ldots
\glt `Yesterday, it was clear that the County official has declared \ldots'
\glend
\item\label{ex:title2} 
\gll Fylkesmannen kj{\o}rte i gr{\o}fta s{\o}r for Oslo.
County-man drove in ditch south of Oslo
\glt `The county official drove into a ditch south of Oslo'
\glend
\end{examples}

\paragraph{Hyphenation} 
Names that include hyphens should be annotated as a named entity if and only if they constitute a new name, e.g. \norex{Lillehammer-saken}$_{EVT}$ \eng{the Lillehammer case} which denotes a specific court case, hence has a unique reference. The similar, but generic, \norex{Lillehammer-advokaten} \eng{the Lillehammer lawyer} is not a named entity, but rather a compound noun, and should therefore receive the special category \etype{DRV}.

\subsubsection{Ambiguity and Metonymy}
Ambiguity is a frequent source of doubt when annotating. This is often caused by so-called metonymical usage, where an entity is referred to by the name of another, closely related entity \cite{Lak:Joh:80}. 
In the annotation of the NorNE corpus we have tried to resolve the ambiguity and choose the entity type based on the context (the document). 
We assume that every entity has a base, or literal, meaning and that when there is ambiguity, either genuinely or due to a lack of context, we resort to the literal meaning of the word(s) \cite{Mar:Nis:02}. For instance, in the example in (\ref{ex:ambig}) below, the context does not clearly indicate whether this is a reference to a geo-political location or organization. We here assume that the location sense is the literal sense of the word \norex{Vietnam} and that the organization sense is by metonymical usage, hence the annotation is \etype{GPE$\_$LOC}.
\begin{examples}
\item\label{ex:ambig}
\gll \underline{Vietnam}$_{GPE\_LOC}$ er flott.
Vietnam is great.
\glt `Vietnam is great.'
\glend
\end{examples}

\subsection{Annotation Process} 
\label{sec:process}
The annotation of the NorNE corpus was performed by two trained linguists, and all documents in the Bokmål section were doubly annotated. As the second phase of the project, the Nynorsk section was annotated by a single annotator. Disagreements in annotations were discussed jointly at regular intervals and the final analysis was agreed upon by both annotators, often followed by an update to the annotation guidelines. 
All annotation was performed using the Brat web-based annotation tool \cite{Ste:Pyy:Top:14}. 

\subsection{Inter-Annotator Agreement}
\label{sec:iaa}

The annotation proceeded in two stages; a first round was used to train the annotators, and hence is not representative of the subsequent annotation process. We only consider the second round of annotation to calculate inter-annotator agreement scores, prior to consolidation. At this stage, the annotation agreement shows a micro F$_1$-score of 91.5 over a set of 138 documents (of 460 in total), comprising 185k tokens and 8434 entities.

The inter-annotator agreement scores are calculated at the entity-level and the annotation for an entity is considered correct only when: (i) both annotators agree on the exact span of the entity, and (ii) both annotators agree on the entity type.  
Tokens that are not part of an entity, i.e., where both annotators agree on the \etype{O} (outside) tag are not considered. In cases where only one of the two annotators has marked an entity, this is still considered as a disagreement.

\begin{figure}
  \includegraphics[width=0.45\textwidth]{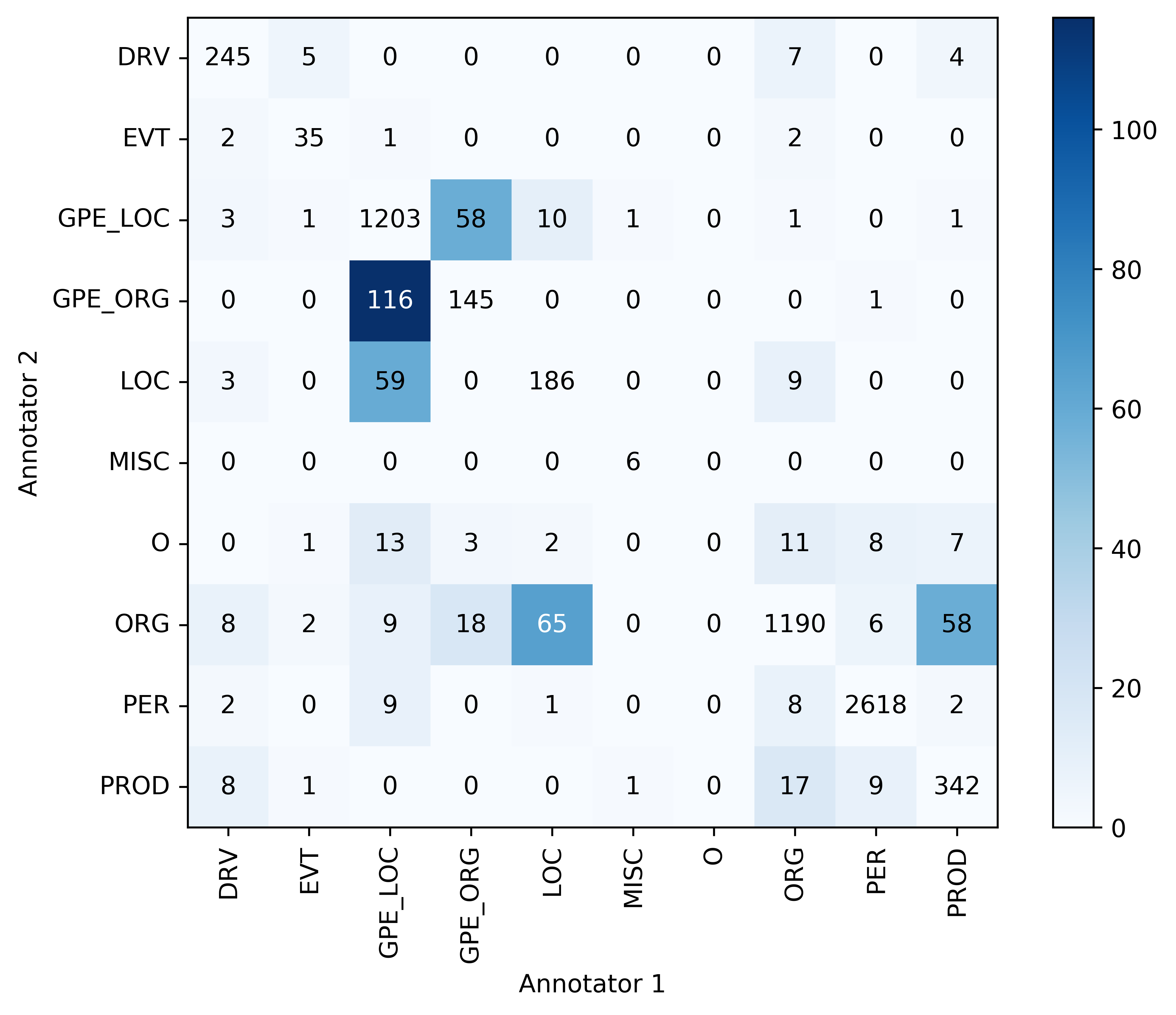}
\caption{Entity confusion matrix for the two annotators.}
\label{fig:iaa}
\end{figure}

Figure~\ref{fig:iaa} shows a confusion matrix for the annotations performed in the second round. While the agreement is generally very high, we observe that the main difficulties of the annotators revolve around the distinction of the subcategories of the \etype{GPE} entity type, in particular confusion between the \etype{GPE\_ORG} and \etype{GPE\_LOC} types themselves. The annotators also have some difficulties in distinguishing these from the main category of \etype{LOC} and in distinguishing between the general \etype{LOC} and \etype{ORG} entity types. Further, we find that the \etype{PROD} category is often confused for the \etype{ORG} category. Table \ref{tab:disagreements} presents some examples of common annotator disagreements. Common entity sub-types among the \etype{PROD}/\etype{ORG} disagreements are newspapers, magazines, web sites and bands.

\begin{table}[t!]
 \centering
 \begin{tabular}{@{}lrr@{}}
  \toprule
  Entity& Annotator 1 & Annotator 2\\ % & \textbf{File:Span}\\
  \midrule
  Afghanistan & \etype{GPE\_LOC} & \etype{GPE\_ORG}\\ % & $db004_0001.ann:711-722$\\
  Norge & \etype{GPE\_ORG} & \etype{GPE\_LOC}\\ % & ap013_0000.ann:4097-4102
  Dagbladet & \etype{ORG} & \etype{PROD}\\ % & $db010_0000.ann:8939-8948$\\
  The Economist & \etype{ORG} & \etype{PROD}\\ % & $ap006_0002.ann:837-850$\\
  Facebook & \etype{ORG} & \etype{PROD}\\ %& blogg-bm002_0001.ann:5379-5387
  Pantera & \etype{PROD} & \etype{ORG}\\ % & $bt001_0008.ann:2028-2035$\\
  \bottomrule
  \end{tabular}
  \caption{Examples of common annotator disagreements.}
  \label{tab:disagreements}
\end{table}

\section{Dataset Overview}
\label{sec:dataset}

The corpus is divided in two parts, one for each of the official written standards of the Norwegian language: Bokmål (\nob) and Nynorsk (\nno). Note that the two parts contain different texts, not translations.  
Global counts of sentences, tokens and annotated entities are shown in Table~\ref{tab:norne}. On a more granular level, 
Tables~\ref{tab:norne-nob} and \ref{tab:norne-nno} summarizes the number of annotations of each entity type for Bokmål and Nynorsk respectively, broken down across the data splits for training, development, and held-out testing. Note that NorNE reuses the same  80-10-10 split previously defined for NDT for both the Bokmål part \cite{HohOvrVel17} and the Nynorsk part \cite{VelOvrHoh17}, which aimed to preserve contiguous texts in the various sections while also keeping the splits balanced in terms of genre. 
Note that, as distributed, NorNE follows the CONLL-U format, with entities labeled according to the IOB2 scheme.

\begin{table}
 \centering
 \begin{tabular}{@{}lccc@{}}
 \toprule
  Standard & Sentences & Tokens & Entities \\ 
  \midrule
  Bokmål (\nob) & $16,309$ & $301,897$ & $14,369$ \\
  Nynorsk (\nno) & $14,878$ & $292,315$ & $13,912$\\
 \bottomrule 
 \end{tabular}
 \caption{Total number of sentences, tokens and annotated named entities in NorNE. The dataset has a separate section for each of the two official written standards of the Norwegian language; Bokmål and Nynorsk.}
\label{tab:norne}
\end{table}

% fixme: are these counts type-level or token-level?
\begin{table}[t!]
 \centering
  \begin{tabular}{@{}lrrrrr@{}}
 \toprule
 Type      & Train & Dev & Test & Total &\%\\
 \midrule
    \etype{PER}     &      4033      &     607      &      560      &      5200      &  36.18    \\
    \etype{ORG}     &      2828      &     400      &      283      &      3511      &  24.43    \\
    \etype{GPE\_LOC}&      2132      &     258      &      257      &      2647      &  18.42    \\
    \etype{PROD}    &      671       &     162      &      71       &      904       &   6.29    \\
    \etype{LOC}     &      613       &     109      &      103      &      825       &   5.74    \\
    \etype{GPE\_ORG}&      388       &      55      &      50       &      493       &   3.43    \\
    \etype{DRV}     &      519       &      76      &      48       &      644       &   4.48    \\
    \etype{EVT}     &      131       &      9       &       5       &      145       &   1.00    \\
\bottomrule
\end{tabular}
\caption{Entity distributions for Bokmål (\nob) in NorNE.}
\label{tab:norne-nob}
\end{table}

\begin{table}[t!]
 \centering
  \begin{tabular}{@{}lrrrrr@{}}
  \toprule
  Type     & Train & Dev& Test & Total &\%\\
  \midrule
    \etype{PER}     &      4250      &     481      &      397      &      5128      &       36.86\\
    \etype{ORG}     &      2752      &     284      &      236      &      3272      &       23.51\\
    \etype{GPE\_LOC}&      2086      &     195      &      171      &      2452      &       17.62\\
    \etype{PROD}    &      728       &     86       &      60       &       874      &        6.28\\
    \etype{LOC}     &      893       &     85       &      82       &      1060      &        7.61\\
    \etype{GPE\_ORG}&      367       &      66      &      11       &      444       &        3.19\\
    \etype{DRV}     &      445       &      50      &      30       &      525       &        3.77\\
    \etype{EVT}     &      141       &      7       &       9       &      157       &        1.12\\
  \bottomrule
  \end{tabular}
  \caption{Entity distributions for Nynorsk (\nno) in NorNE.}
  \label{tab:norne-nno}
\end{table}

\section{Experimental Results and Analysis}
\label{sec:experiments}

In this section we present some preliminary experimental results for named entity recognition using NorNE.  
We investigate the effects of using different mappings of the label set, different label encodings (IOB2, etc), different embedding dimensionalities, as well as joint modeling of the Bokmål and Nynorsk variants. Apart from the joint modeling, the other experiments will target only the Bokmål section of the dataset. Before moving on to the results, we first briefly outline the experimental setup.  

\subsection{Experimental Setup}
\label{sec:setup}

The modeling is performed using NCRF++ \cite{Yan:Zha:18} -- a configurable sequence labeling toolkit built upon PyTorch. 
Following \newcite{Yan:Lie:Zha:18}, our particular model configuration is similar to the architecture of \newcite{Chi:Nic:16} and \newcite{Lam:Bal:Sub:16}, achieving results that are close to state-of-the-art for English on the CoNLL-2003 dataset: it combines a character-level CNN and a word-level BiLSTM, finally feeding into a CRF inference layer. The input to the word-level BiLSTM is provided by the concatenation of (1) the character sequence representations from the CNN using max-pooling in addition and (2)  pre-trained word embeddings from the NLPL vector  repository\footnote{\url{http://vectors.nlpl.eu/repository/}} \cite{Far:Kut:Oep:17}. Further details about the latter are provided in the next section. 

Across all experiments we fix and re-use the same random seed for initializing the models, as to reduce the effect of non-determinism, and otherwise fix the parameters to their default values.\footnote{Parameter settings include the following: 
%word-level BiLSTM = 1 layer of 200 units,  
% character-level units = 50 
%character embedding dimensionality = 30 
optimizer=SGD, 
epochs=50, 
batch size=10,  
dropout=0.50, 
learning rate = 0.015 with a decay of 0.05, 
L2-norm=$1^{-8}$, 
seed=42.}

For model evaluation we follow the scheme defined by the SemEval 2013 task 9.1 \cite{Seg:Mar:Her:13}, using the re-implementation offered by David S. Batista.\footnote{\url{https://github.com/davidsbatista/NER-Evaluation}} 
We report 
F1 for exact match on the entity level, i.e., both the predicted boundary and entity label must be correct.  (This measure was dubbed \textit{strict} in SemEval 2013 task 9.1.)

\subsection{The Use of Pre-Training}
\label{sec:pre-training}

In a preliminary round of experiments, we evaluated the impact of pre-training on the model. The word embeddings are trained on the Norwegian News Corpus (over 1 billion tokens of Norwegian Bokmål) 
%  and 60 million tokens of Norwegian Nynorsk 
and NoWaC (Norwegian Web as Corpus; approximately 687 million tokens of Bokmål) 
using fastText  
\cite{Boj:Gra:Jou:17} with a vocabulary of 2.5 million unique words and a window size of 5. We here re-use embeddings that are made available by the NLPL vector repository \cite{Far:Kut:Oep:17}; for more details on the training process see \newcite{Stadsnes:18} and \newcite{Sta:Ovr:Vel:18}. 

Table~\ref{tab:pretraining} shows results for the Bokmål development split both with and without pre-training, and using both the CBOW and SkipGram algorithm as implemented in fastText. Unsurprisingly, we observe that pre-training substantially boosts performance of the NER model. Moreover, we observe that CBOW (here only shown for a 100-dimensional model) is substantially outperformed by the SkipGram model, and that performance steadily increase with increased dimensionality. In all the subsequent  experiments reported in the paper we use a SkipGram model with a dimensionality of 600. 

\begin{table}[t!]
	\centering
	\begin{tabular}{@{}crr@{}}
		\toprule
		Pre-train. & Dim. & F$_1$ \\ 
		\midrule
  none & 100 & 76.54 \\ 
  CBOW & 100 & 84.36 \\ 
	SG &  50 & 87.61 \\
	SG & 100 & 89.47 \\
	SG & 300 & 90.02 \\
	SG & 600 &\bf{90.75} \\ 
		\bottomrule
	\end{tabular}
	\caption{Evaluating the impact of fastText pre-training, testing on the Bokmål development split of NorNE.}
	\label{tab:pretraining}
\end{table}

\subsection{Label set and Label Encoding}
\label{sec:encoding}

In this section we investigate the interactions between choice of label set and label encoding. On the one hand we experiment with the granularity of the label set or entity types; mapping the original entity types to a smaller set of more general types. On the other hand we experiment with mapping the IOB labels specified in the distributed corpus to variations of the BIOES (BIOLU) label encoding scheme. 

\subsubsection{Label Set} 
%\paragraph{Label set} 
We consider the following label mappings: 
\begin{list}{$\circ$}{} 
	\item \textbf{NorNE-full}: Using the full set of 8 entity types, as in the experiments above. 
	\item \textbf{NorNE-7}: Conflating instances of the geo-political subcategories \etype{GPE\_ORG} and \etype{GPE\_LOC} to the more general type \etype{GPE}, yielding 7 entity categories.
	\item \textbf{NorNE-6}: Dispensing with the geo-political types entirely, merging \etype{GPE\_ORG} and \etype{GPE\_LOC} into \etype{ORG} and \etype{LOC} respectively, yielding 6 entity categories.
\end{list}

The question of what comprises the most suitable level of granularity ultimately depends on the downstream task, but in this section we report experimental results for training and testing with the different labels sets to analyze the learnability of the different granularities.

\begin{figure}[!hbpt]
	\begin{subfigure}[b]{0.75\linewidth}
		\flushright
		\includegraphics[scale=0.5]{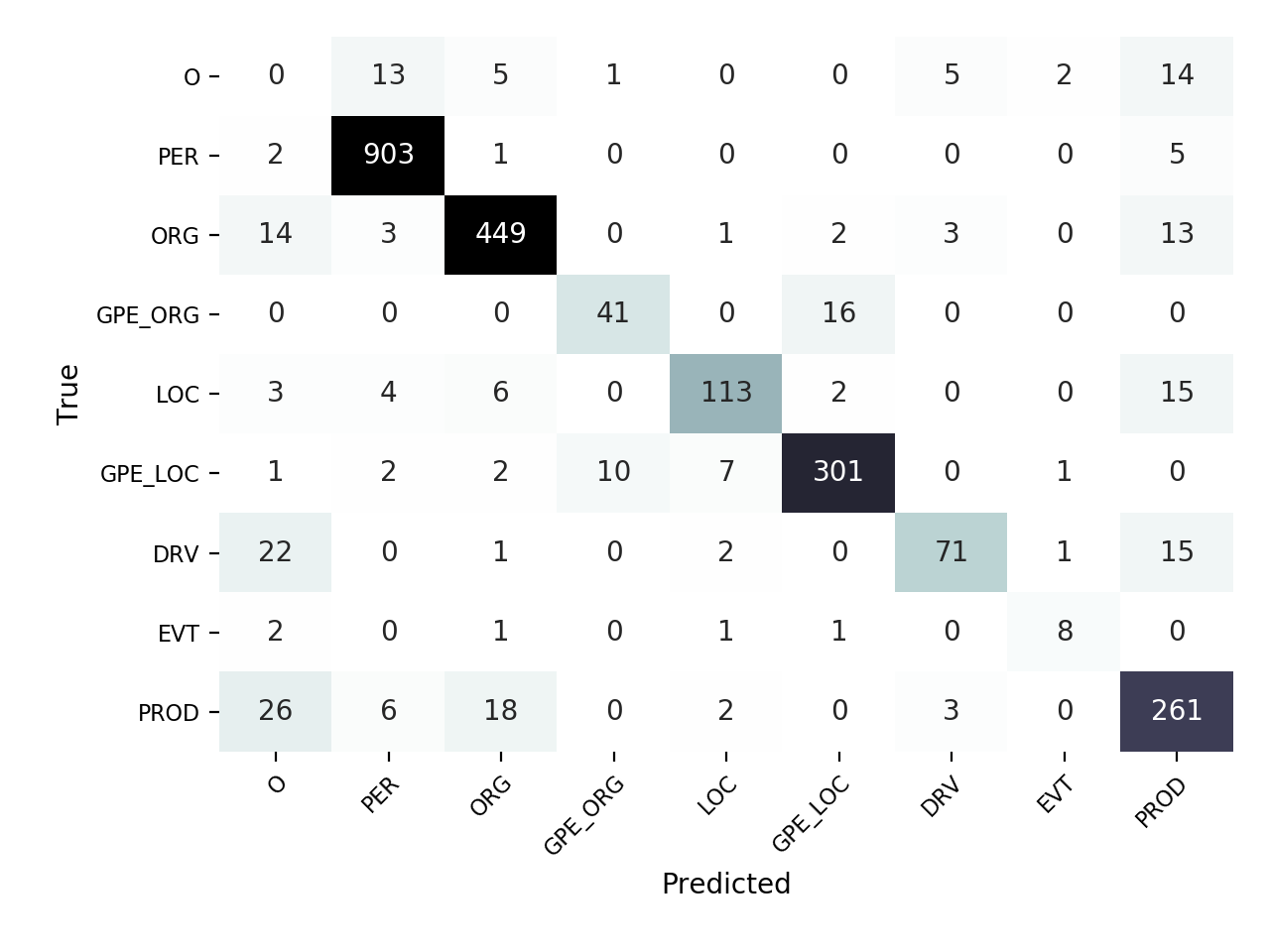}
		\subcaption{NorNE-full}
	\end{subfigure}
	
	\begin{subfigure}[b]{0.75\linewidth}
		\flushright
		\includegraphics[scale=0.5]{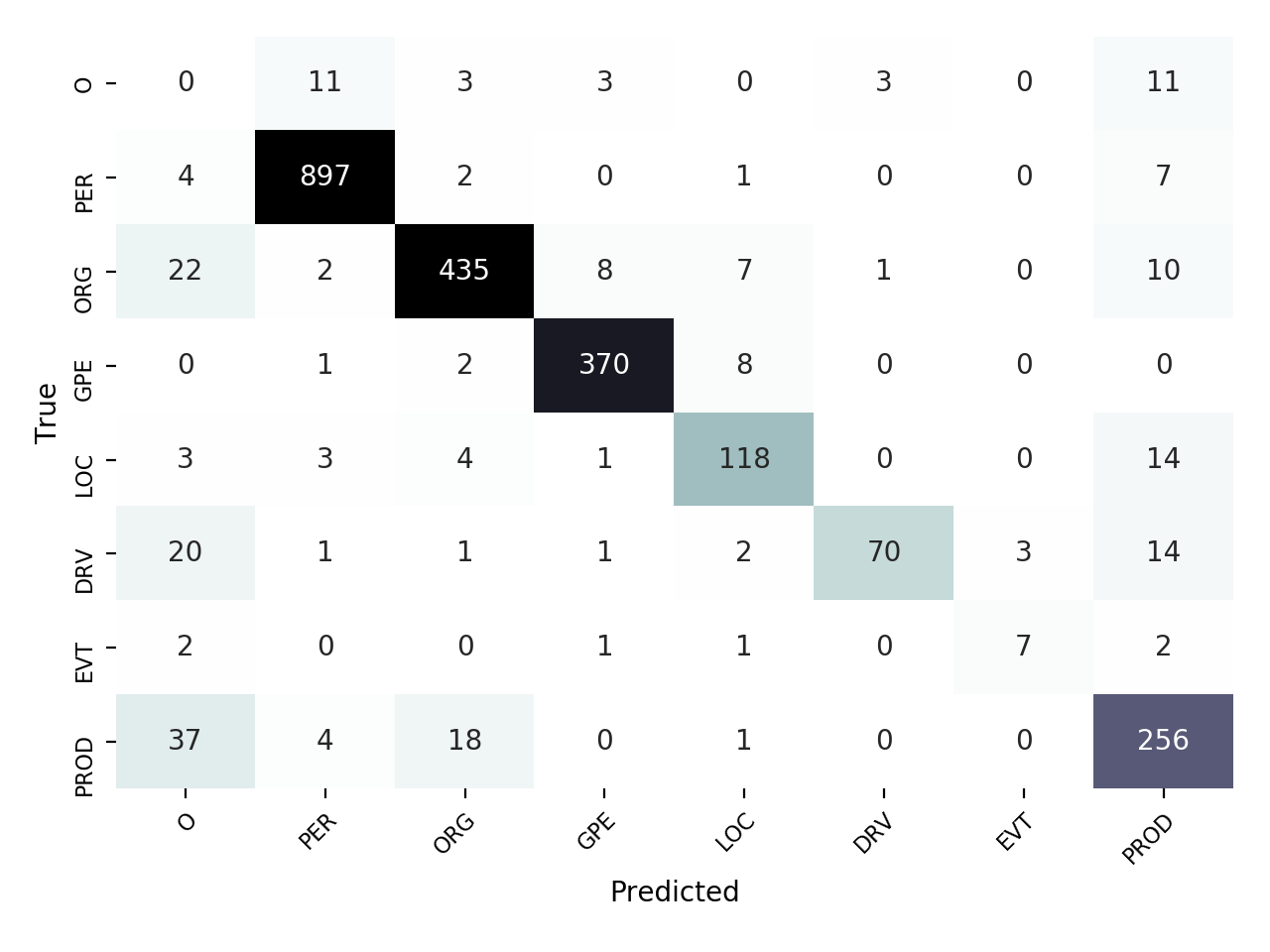}
		\subcaption{NorNE-7}
	\end{subfigure}

	\begin{subfigure}[b]{0.75\linewidth}
		\flushright
		\includegraphics[scale=0.5]{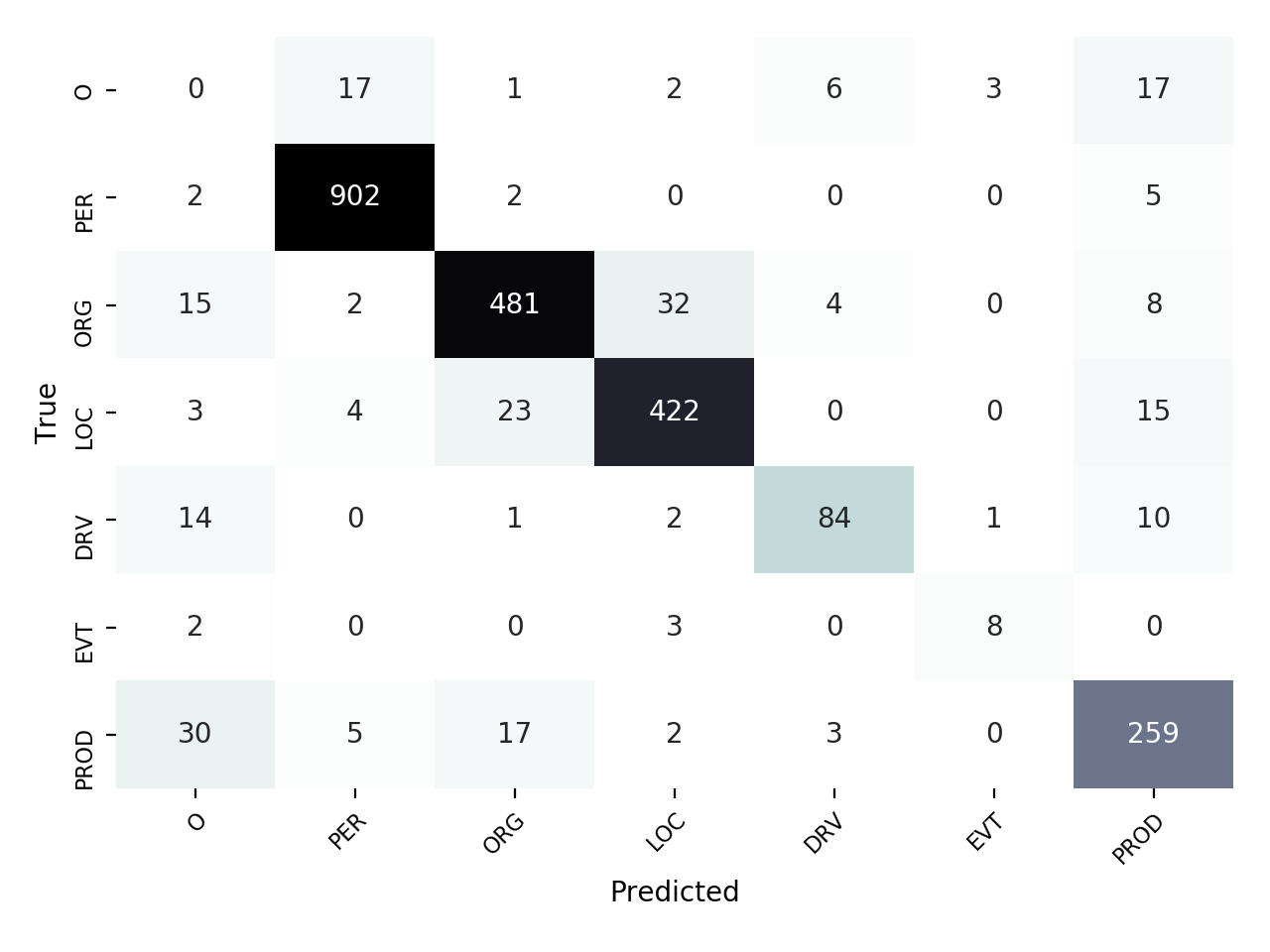}
		\subcaption{NorNE-6}
	\end{subfigure}
	\caption{Aggregated confusion matrices for the different label granularities; NorNE-full, NorNE-7 and NorNE-6.}
	\label{fig:heatmaps}
\end{figure}

\begin{figure*}[h!]
\centering
    \begin{dependency}
        \begin{deptext}[column sep=.2cm]
             \& \& \& \tag{B-PER} \& \& \& \tag{B-GPE\_LOC} \& \\
    Til \& dagleg \& jobbar \& Sydnes \& mykje \& utafor \& Noregs \& landegrenser \\
    Til \& daglig \& jobber \& Sydnes \& mye \& utenfor \& Norges \& landegrenser \\ 
    {\it To} \& {\it daily} \& {\it works} \& {\it Sydnes} \& {\it lot} \& {\it outside} \& {\it Norway's}  \& {\it borders} \\
        \end{deptext}
    \end{dependency}
    \caption{Example sentence in Nynorsk (second row) and Bokm{\aa}l (third) with English gloss (bottom) and IOB  named entity labels (top). A fluent translation to English would be \textit{On a daily basis Sydnes works a lot outside of Norway's national borders.}}
 \label{fig:parallel}
\end{figure*}

\subsubsection{Label Encoding}
The annotations of NorNE are distributed using the standard IOB(2)  scheme.\footnote{In IOB2, the B-label is used at the beginning of every named entity, regardless of its span, while in the IOB1 variant the B-label is only used when a named entity token is followed by a token belonging to the same entity.} However, this can be easily mapped to other variations like IOBES, where the extra E-label indicates the end-token of an entity and the S-label indicates single, unit-length entities. (This latter scheme also goes by other names like BIOLU.) Several studies have reported slight performance increases when using the  IOBES encoding compared to IOB  \cite{Rat:Rot:09,Yan:Lie:Zha:18,Rei:Gur:17}. However, it is typically not clear whether the benefits stem from adding the E- or S-labels or both.

\subsubsection{Results}

Table~\ref{tab:encoding} report experimental results for all of these variations -- i.e. isolating the effects of the E- and S-labels -- and across all the three different sets of entity types discussed above. 

\begin{table}[t!]
 \centering
 \begin{tabular}{@{}lcccc@{}}
 \toprule
 %\textbf{Label set} 
  & IOB &  IOBE & IOBS & IOBES \\ 
 \midrule
 NorNE-full & 90.75 & 90.45  & \textbf{90.76} &  90.58 \\
 \addlinespace
 NorNE-7    & 91.86 & 91.80 &  91.99 &  \textbf{92.29}  \\
 \addlinespace
 NorNE-6    & 90.95 & 90.50 & \textbf{91.85} &  91.38 \\
 \bottomrule
 \end{tabular}
 \caption{Perfomance comparison of models trained on NorNE-full using IOB, IOBE, IOBS and IOBES encodings, when evaluated on the NorNE development set, using label sets of different granularities.}
 \label{tab:encoding}
\end{table}

There are several things to notice here. IOBE seems to have a negative performance impact regardless of the chosen label set. Also, compared to the standard IOB  encoding, IOBES also has a negative impact paired with NorNE-full, but gives improved results together with NorNE-7 and NorNE-6.  It is not easy to pinpoint \emph{exactly} why we get lower results for NorNE-full when used with more fine-grained encodings, but possible explanation could be that the resulting increased class sparsity has more of an effect using the full label-set.  

Interestingly, we find that IOBS gives better results for all label sets, giving the highest F1 scores for both NorNE-full and NorNE-6, and also a marginal performance increase for NorNE-7.
In other words, no matter the label set it seems beneficial to use a dedicated tag for single-token entities, while the  benefits of including an end-tag (or both) are less clear. 

Regardless of the chosen encoding, we see that the NorNE-7 label-set yields the highest scores. We also see that reducing the label granularity always leads to higher absolute scores compared to using the full label-set. This is not in itself informative however and is an effect that be expected just from the fact the label ambiguity is reduced.

Figure~\ref{fig:heatmaps} shows confusion matrices for all label-sets, using models trained with the IOB encoding, making for some interesting observations.
First of all, we see that there is little confusion among the entities
\etype{ORG}, \etype{GPE\_ORG}, \etype{LOC} and \etype{GPE\_LOC} in
NorNE-full. At the same time, collapsing \etype{GPE\_LOC} and \etype{GPE\_ORG} to a single category in NorNE-7 does not seem detrimental, with a marginal amount of confusion between \etype{LOC}, \etype{ORG}, and \etype{GPE}. However, with NorNE-6, valuable information appears to have been lost when removing the geo-political category, introducing more confusion between the location and organization category.

In general, we see that products (\etype{PROD}) seems to be a difficult category to
classify, likely reflecting the rather heterogeneous character of this
category. Moreover, this entity category has the longest average token span, and
one might suspect that long entities might have more boundary-related errors,
which could explain the high confusion with the \textit{outside-of-entity}
class.

Of course, the appropriate choice of label set ultimately depends on the
downstream use case. However, unless one really needs to distinguish between
the \etype{GPE} sub-categories, our experiments above seem to point to NorNE-7
label set as a good option, possibly in combination with an IOBES encoding.

\subsection{Joint Modeling of Bokmål and Nynorsk}
\label{sec:joint}

As mentioned in the introduction, there are two official  written  standards  of the Norwegian language; Bokmål  (literally `booktongue') and Nynorsk (literally `new Norwegian'). Focusing on dependency parsing, \newcite{VelOvrHoh17} investigated the interactions across the two official standards with respect to parser performance. The study demonstrated that while applying parsing models across standards yields poor performance, combining the training data for both standards yields better results than previously achieved for each of them in isolation. We here aim to investigate similar effects for named entity recognition.

\subsubsection{Background: On Nynorsk and Bokmål}  
While Bokmål is the main variety,  roughly 15\% of the Norwegian population uses Nynorsk. However, language legislation specifies that minimally 25\%  of  the written public service information should be in Nynorsk. The same minimum ratio applies to the programming of the Norwegian Public Broadcasting Corporation (NRK). The two varieties are so closely related that they may in practice be regarded as `written dialects'. However, lexically  there  can  be  relatively  large differences.   

Figure~\ref{fig:parallel} shows an example sentence in both Bokm{\aa}l and Nynorsk.  While the word order is identical and many of the words are clearly related, we see that only 3 out of 8 word forms are identical. When quantifying the degree of lexical overlap with respect to NDT -- the treebank data that we too will be using -- \newcite{VelOvrHoh17} find that out of the 6741 non-punctuation word forms in the Nynorsk development set, 4152, or 61.6\%, of these are unknown when measured against the Bokm{\aa}l training set. For comparison, the corresponding proportion of unknown word forms in the Bokm{\aa}l development set is 36.3\%. These lexical differences are largely caused by differences in productive inflectional forms, as well as highly frequent functional words like pronouns and determiners.

\subsubsection{A Joint NER Model}  
For the purpose of training a joint NER model we also train a new version of the fastText SkipGram embeddings on the NNC and NoWaC corpora, using the same parameters as before (and 600 dimensions), but this time including the available Nynorsk data for NNC, amounting to roughly 60 million additional tokens.

Several interesting effects can be observed from the results in Table~\ref{tab:joint}. The first two rows show the results of training single-standard NER models (like before) with the joint embedding model, but this time also testing across standards; training a model on Bokmål and applying it to Nynorsk, or \textit{vice versa}. As can be clearly seen, performance drops sharply in the cross-standard settings (italicized in the table). For example, while the Bokmål NER model achives an F1 of 89.47 on the Bokmål development data, the performance plummets to 82.34 when the same model is applied to Nynorsk.

The last row of Table~\ref{tab:joint} shows the effects of training a \textit{joint} NER model on the combination of the Bokmål and Nynorsk data (randomly shuffling the sentences in the combined training and validation splits).  
% fixme: bør vi utdype at singe-standard modellene bruker single-standard validation?
We see that the joint model substantially outperforms both of the single-standard models on their respective development splits. On the heldout splits, the joint model again has much better performance for Nynorsk, although the single-standard setup shows slightly better results for Bokmål.  

The results for the joint modeling setup is a double-win with immediate practical consequences: Not only do we see comparable or increased performance, it also means we only need to maintain a single model when performing NER for Norwegian. The alternative would be to either accept a sharp drop in performance whenever Nynorsk, say, was encountered, \textit{or} to first perform language identification to detect the given variety and then apply the appropriate model.  

\begin{table}[t!]
 \centering
 \begin{tabular}{@{}lccccc@{}}
  \toprule
  &   \multicolumn{2}{c}{Development} 
  & & \multicolumn{2}{c}{Heldout} \\ 
  \cmidrule{2-3} \cmidrule{5-6}
  Training   & \nob    & \nno    & & \nob    & \nno \\ 
  \midrule
  \nob         &    89.47 &\it{82.34}& &\bf{83.89}&\it{81.59}\\
  \nno         &\it{84.01}&    86.53 & &\it{76.88}&    83.89 \\
  \nob+\nno    &\bf{90.92}&\bf{88.03}& &    83.48 &\bf{85.32}\\
  \bottomrule
  \end{tabular}
  \caption{Joint and cross-standard training and testing of NER models. The first column indicates the language standard used for training the NER model; either Bokmål (\nob), Nynorsk (\nno), or both. The Development and Heldout columns shows F1 scores when testing on the respective splits of either standard. The italicized scores correspond to a cross-standard setup where the language variety used for training is different from testing. Bold indicates best performance.}
  \label{tab:joint}
\end{table}

\section{Summary}
\label{sec:summary}

This paper has documented a large-scale annotation effort adding named entities to the Norwegian Dependency Treebank. The resulting dataset -- dubbed NorNE -- is the first publicly available\footnote{\url{https://github.com/ltgoslo/norne/}} dataset for named entity recognition (NER) for Norwegian and covers both of the official written standards of the Norwegian language -- Bokmål and Nynorsk -- comprising roughly 300,000 tokens of each. The annotations include a rich set of entity types including persons, organizations, locations, geo-political entities, products, and events, in addition to a class corresponding to nominals derived from names. In addition to discussing the principles underlying the manual annotations, we provide an in-depth analysis of the new dataset through an extensive series of first benchmark NER experiments using a neural sequence labeling architecture (combining a character-level CNN and a word-level BiLSTM with a CRF inference layer). Among other results we demonstrate that it is possible to train a joint model for recognizing named entities in Nynorsk and Bokmål, eliminating the need for maintaining separate models for the two language varieties.

\section*{Acknowledgements} 
NorNE is the result of a  collaborative effort involving the Schibsted Media Group, Spr{\aa}kbanken (`the Language Bank') of the National Library of Norway, and the Language Technology Group at the Department of Informatics, University of Oslo. We would like to thank the annotators, Vilde Reichelt and Cato Dahl, for their meticulous efforts in providing accurate and consistent annotations.

\section*{Bibliographical References}

%\bibliographystyle{lrec}
%\bibliography{norne.bib}

\end{document}